\documentclass[10pt,twocolumn,letterpaper]{article}

\usepackage{cvpr}              
\definecolor{cvprblue}{rgb}{0.21,0.49,0.74}
\usepackage[pagebackref,breaklinks,colorlinks,allcolors=cvprblue]{hyperref}

\usepackage{url}
\usepackage{graphicx}
\usepackage{multirow}
\usepackage[table]{xcolor}
\usepackage{caption}
\usepackage{subcaption}
\usepackage{pifont}

\usepackage{amsmath}
\usepackage{amssymb}

\usepackage[export]{adjustbox}

\def\paperID{18289} 
\def\confName{CVPR}
\def\confYear{2026}

\title{Vision Remember: Recovering Visual Information in Efficient LVLM with Vision Feature Resampling}

\author{Ze Feng\\
Southeast University\\
\and
Jiang-jiang Liu\\
Baidu VIS\\
\and
Sen Yang\\
Baidu VIS\\
\and
Lingyu Xiao\\
The University of Hong Kong\\
\and
Zhibin Quan\\
Southeast University\\
\and
Zhenhua Feng\\
Jiangnan University\\
\and
Wankou Yang \\
Southeast University\\
\and
Jingdong Wang\\
Baidu VIS\\
}

\begin{document}
\maketitle
\begin{abstract}
The computational expense of redundant vision tokens in Large Vision-Language Models (LVLMs) has led many existing methods to compress them via a vision projector.
However, this compression may lose visual information that is crucial for tasks relying on fine-grained spatial relationships, such as OCR and Chart \& Table Understanding.
In this paper, we propose to resample original vision features across the LLM decoder layers to recover visual information and attain efficiency. 
Following this principle, we introduce Vision Remember, which includes two key modules: (1) Token-Feature Cross-Attention Layer and (2) Token Bidirectional Self-Attention Layer.
In the Token bidirectional attention, we employ self-attention mechanism to maintain the bidirectional interaction between vision tokens and the text-guided token.
In the Token-Feature interaction attention, we introduce local cross-attention to resample the visual feature and utilize the multi-level fusion to enrich the visual representation.
We conduct comprehensive experiments on multiple visual understanding benchmarks and the results with the LLaVA-NeXT baseline show that Vision Remember outperforms TokenPacker by +2.7 and FastV by +5.7 across nearly all the settings.
Compared with previous vision feature re-fusion methods, our approach also surpasses DeepStack by +3.9 and SVA Aggregator by +3.4 on the same baseline.
The experimental results validate the generalization capability of the proposed method when combined with various efficient vision projectors and LVLMs.
\end{abstract}

\section{Introduction}
\label{sec:intro}

In recent years, with the rapid advancement of Large Language Models (LLMs)~\citep{achiam2023gpt,touvron2023llama,bai2023qwen,cai2024internlm2,liu2024deepseek}, a growing body of research has focused on integrating visual parsing, understanding and generation capabilities into LLM, leading to the development of a series of Large Vision-Language Models (LVLMs)~\citep{alayrac2022flamingo,bai2023qwenvl,li2023blip,liu2024improved,liu2024visual,lu2024deepseekvl}.
The general approach involves aligning vision tokens with the linguistic domain via a projector and then concatenating with text tokens before feeding them into an LLM.

However, vision encoders often produce a large number of vision tokens (e.g., 576 in LLaVA-1.5~\citep{liu2024improved}, max 2880 in LLaVA-NeXT~\citep{liu2024llavanext}, and max 5760 in LLaVA-OneVision~\citep{li2024llavaov} for an image), which occupy the majority of the input embeddings. 
Due to the quadratic complexity of the attention mechanism with respect to the number of tokens, longer input embeddings consume significant computational resources and memory, impeding the applications of LVLMs in practice, particularly under computationally constrained scenarios such as edge computing and robotics.

Many existing studies try to improve the efficiency and have found that vision tokens exhibit significant redundancy~\citep{chen2024image,zhang2024sparsevlm,xing2024pyramiddrop}. 
As a result, they have made efforts to reduce the number of vision tokens. 
There are two typical approaches: (1) redesigning the projector to directly compress the vision tokens~\citep{yao2024deco,cha2024honeybee,chu2023mobilevlm,chu2024mobilevlm,li2025tokenpacker,shen2024mome}, and (2) pruning unimportant vision tokens~\citep{chen2024image,xing2024pyramiddrop,zhang2024sparsevlm}. 
For example, DeCo~\citep{yao2024deco} employs Adaptive Average Pooling, and Qwen2.5-VL~\citep{bai2025qwen2} uses PixelShuffle to compress vision tokens.
VisPruner~\citep{vispruner} maintains the dominant tokens and prunes the other tokens in the vision encoder.
Nonetheless, these methods may lose visual information, which is important for the tasks that rely on fine-grained spatial relationships, such as OCR and Chart \& Table understanding.

\begin{figure}[tbp]
  \centering
  \begin{subfigure}[t]{0.45\textwidth} 
    \includegraphics[width = \linewidth, valign = t]{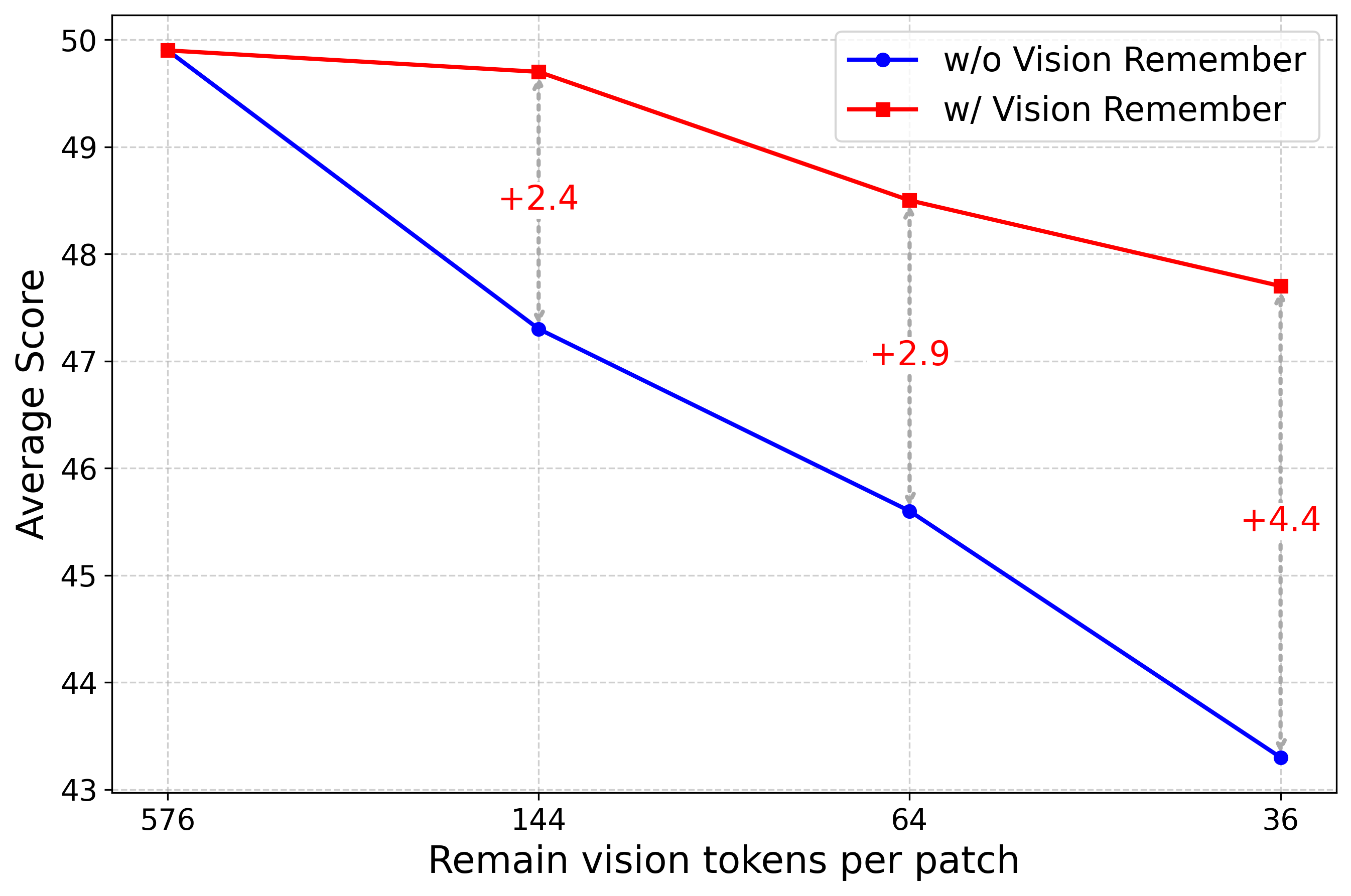} 
    \caption{Performance drop with compression ratios.}
    \label{fig:performance}
  \end{subfigure}
  \hfill 
  \begin{subfigure}[t]{0.473\textwidth} 
    \includegraphics[width = \linewidth, valign = t]{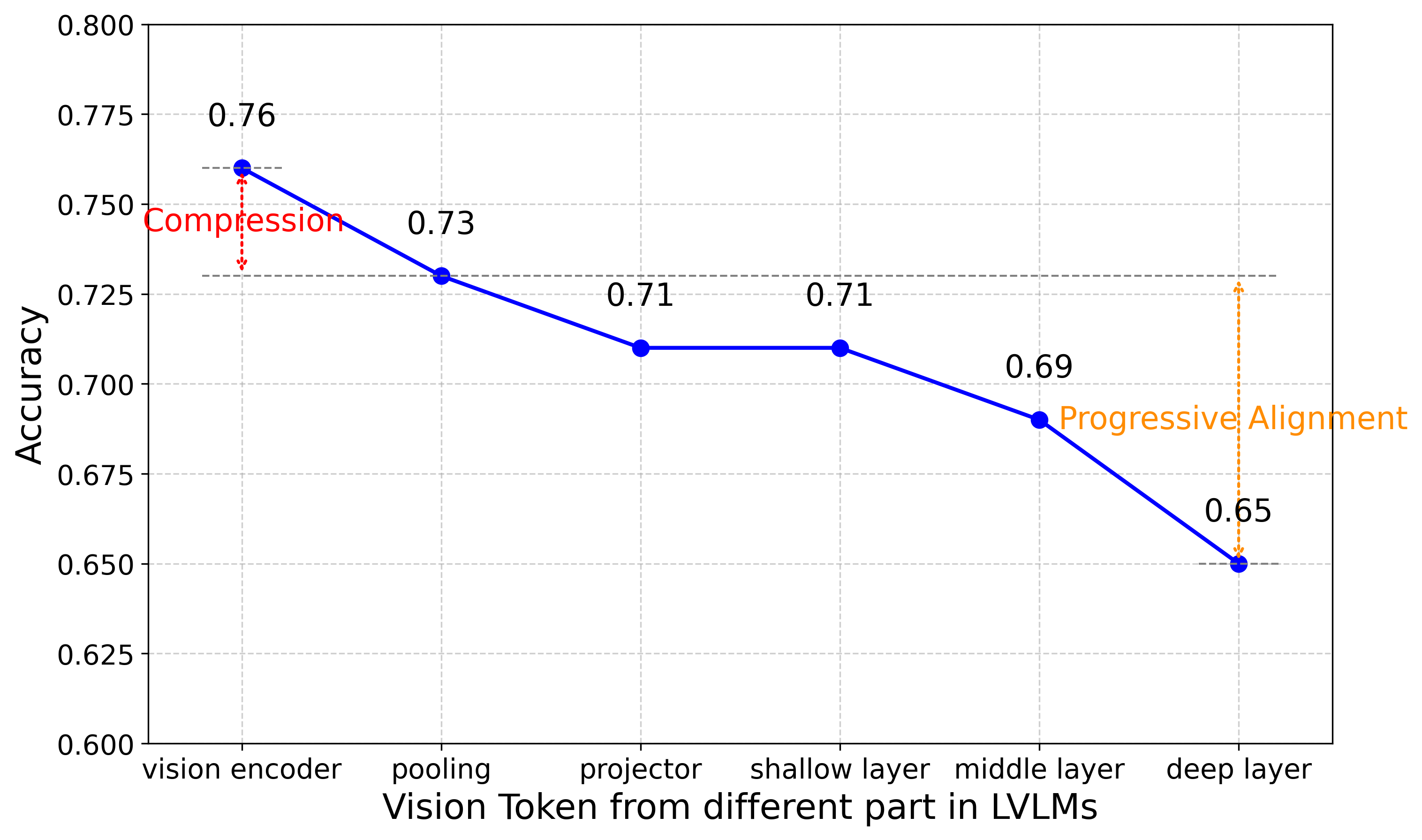}
    \caption{Linear probing of classification on Tiny-ImageNet}
    \label{fig:linear_probing}
  \end{subfigure}
  \caption{Preliminary analysis. (a) Compressing the vision tokens can cause information loss, resulting in performance degradation. The proposed Vision Remember alleviates this problem. (b) We extract vision tokens from distinct components of LVLM and evaluate the classification accuracy on Tiny-ImageNet. The compression only happens in pooling. Our analysis identifies two primary sources of visual information loss: \textbf{Information Bottleneck} in Token Compression and \textbf{Visual Cues Forgetting} in Progressive Alignment.}
  \label{fig:vs}
\end{figure}

To systematically identify the reasons for visual information loss, we evaluate the classification capacity of vision tokens extracted from different components of LVLM on the TinyImageNet dataset.
Similar to linear probing, we freeze all the parameters in LVLM and only train a lightweight classification head composed of a single cross-attention layer followed by a linear layer.
As shown in Fig.~\ref{fig:linear_probing}, we identify two fundamental reasons for the performance degradation: 
(1) Information Bottleneck in Token Compression - The compression of vision tokens inevitably discards fine-grained visual details (e.g., texture patterns, small objects), while the surviving tokens lack the representational capacity to reconstruct such high-frequency visual information; 
(2) Visual Cues Forgetting in Progressive Alignment - During the cross-modal alignment process, where vision tokens sequentially interact with text tokens in the LLM’s attention layers, visual features undergo gradual attenuation due to dominant linguistic priors, resulting in visual cues forgetting across the LLM decoder.
Hence, we raise a question: \textit{Since the problem comes from the compression in the projector and the forgetting in LLM, can we recover the original vision features between the LVLM decoder layers?}

To answer the above question, this paper presents \textit{\textbf{Vision Remember}}, an approach that resamples original visual features multiple times across the LVLM decoder layers to compensate for the lost vision cues.
The main motivation is that the features obtained by the vision encoder contain original vision information, and we can re-inject them into vision tokens, not in the projector, but between the decoder layers.
Following this principle, we introduce the first key module: \textit{\textbf{Token-Feature Cross-Attention Layer}}, which employs local cross-attention to interact the vision tokens and vision features. 
Furthermore, we also aggregate multi-level features to enrich the visual representation and enhance the model’s ability of visual comprehension.
Another key module is \textit{\textbf{Token Bidirectional Self-Attention Layer}}.
Casual attention mask inherently restricts cross-token interactions in visual representations while preventing access to subsequent textual cues, consequently disregarding textual descriptions of foreground objects.
To address this issue, this module employs self-attention to enable mutual attention among vision tokens, and introduces text-guided tokens to implicitly characterize the region of interest.

We evaluate our method on LLaVA-NeXT~\citep{liu2024llavanext}, the most widely used baseline in academia, and assess the model's performance through average scores across eleven comprehensive benchmarks.
Experimental results demonstrate consistent performance gains when our method is combined with various efficient visual projectors. Specifically, Vision Remember achieves improvements of +3.0 (6.6\%), +3.2 (7.2\%), and +4.4 (10.1\%) for Average Pooling, PixelShuffle, and Perceiver Resampler, respectively. 
On identical baselines, our approach outperforms prior works, TokenPacker \citep{li2025tokenpacker} and FastV \citep{chen2024image}, by margins of +2.7 (5.9\%) and +5.7 (13.3\%).
To further validate the generalizability, we conduct experiments on two different baselines: Qwen2.5-VL \citep{bai2025qwen2} and MiniCPM-V \citep{yao2024minicpm}, and observe performance improvements. 
These experiments demonstrate that Vision Remember can serve as a fundamental component when constructing an efficient LVLM.

\section{Related Work}
\label{sec:related}

\subsection{Large Vision-Language Models}
Many works focus on endowing LLMs with visual understanding capabilities, transforming them into LVLMs \citep{yao2024minicpm,liu2024revisitingmllmsindepthanalysis,chen2024far,liu2024improved,abdin2024phi,liu2024llavanext,cambrian,wang2024qwen2vl,lu2024deepseekvl}.
Based on differences in visual signal integration methods, we categorize existing approaches into two classes: (1) Token Concatenation and (2) Visual Feature Sampling.

\noindent\textbf{Token Concatenation based LVLM.}
These methods align the vision tokens into the linguistic domain by a projector, and then concatenate them with text tokens before feeding them into the LLM.
LLaVA series \citep{liu2024visual, liu2024llavanext, liu2024improved, li2024llavaov} adopt this paradigm and directly employ MLP layers to map vision tokens into the language domain.
Cambrian-1 \citep{cambrian} has explored various combination methods of multiple vision encoders.
DenseConnector \citep{yao2024dense} enhances existing LVLMs by leveraging multi-level visual features.
However, the aforementioned methods primarily focus on enhancing the understanding capabilities of LVLMs, while neglecting the efficiency of the models.

\noindent\textbf{Visual Feature Sampling based LVLM.}
Several approaches inject visual information into LLMs via cross-attention layers, where text tokens serve as queries while visual features act as keys and values.
Flamingo \citep{alayrac2022flamingo} introduced gated x-attention layers, which enable the model to understand visual inputs by employing 
LLaMA 3 \citep{dubey2024llama3} also adopts this paradigm, constructing multimodal models with varying parameter counts, and achieves strong performance through large-scale training.
EVLM \citep{chen2024evlm} and NVLM \citep{dai2024nvlm} integrate these two paradigms, constructing hybrid-architecture LVLMs.
Unlike previous approaches, our method performs sampling exclusively on the vision tokens by leveraging local cross-attention mechanisms. 

\subsection{Efficient Large Vision Language Models}
Many works focus on improving the efficiency of LVLMs by reducing the number of visual tokens, which can generally be categorized into the following two types:
(1) redesigning the projector to directly compress the visual tokens; 
(2) directly pruning the unimportant vision tokens between the decoder layers.

\noindent\textbf{Projector Design.}
DeCo \citep{yao2024deco} proposes using 2D adaptive average pooling directly in the projector to perform downsampling of visual tokens.
By utilizing Point-to-Region attention in the local region, TokenPacker \citep{li2025tokenpacker} enhances fine-grained understanding capability while preserving spatial information.
MobileVLM \citep{chu2023mobilevlm, chu2024mobilevlm} introduces a convolutional LDP module for visual token compression, whereas Qwen2-VL \citep{wang2024qwen2vl} and InternVL \citep{chen2024internvl} employ PixelShuffle.

\noindent\textbf{Vision Token Pruning.}
FastV \citep{chen2024image} introduces a method that prunes the last top-k visual tokens based on attention values. This plug-and-play approach can be integrated into various LVLMs in a training-free paradigm.
PyramidDrop \citep{xing2024pyramiddrop} divides the entire LVLM decoder into multiple stages and performs pruning at a fixed ratio after the last layer in each stage.
VisionZip \citep{yang2025visionzip} and VisPruner \citep{vispruner} prunes the redundant tokens in the vision encoder, while preserving the structural integrity of LLM.

Unlike existing approaches, our proposed method focuses on recovering lost visual cues rather than further optimizing model operational efficiency.

\subsection{Vision Feature Re-Fusion in LVLMs}
Built upon Token Concatenation based LVLMs, some methods try to re-sample and re-fuse the vision feature in the LLM.
FlexAttention \cite{li2024flexattention} proposes a flexible attention mechanism in which the selected high-resolution tokens are concatenated to the low-resolution tokens, then input to a hierarchical self-attention layer.
However, FlexAttention could not be compatible with FlashAttention or SDPA, making the training and inference cannot achieve the obvious acceleration.
DeepStack \cite{liu2024deepseek} proposes to stack the high-resolution token into multiple groups and feed each group to the aligned decoder layer, achieving accuracy improvement while minimal additional cost.
SVA Aggregator proposed in Cambrian-1 \cite{cambrian} uses vision tokens between decoder layers to query the vision features belonging to multiple vision encoders, but ensembling multiple encoders harms the inference efficiency. 
Different from the previous methods, our proposed approach could appropriately leverage the visual representations and alleviate the visual information loss, outperforming them with a large margin.

\section{Vision Remember}
\label{sec:method}

\begin{figure*}[t]
    \centering
    \includegraphics[width=1.0\linewidth]{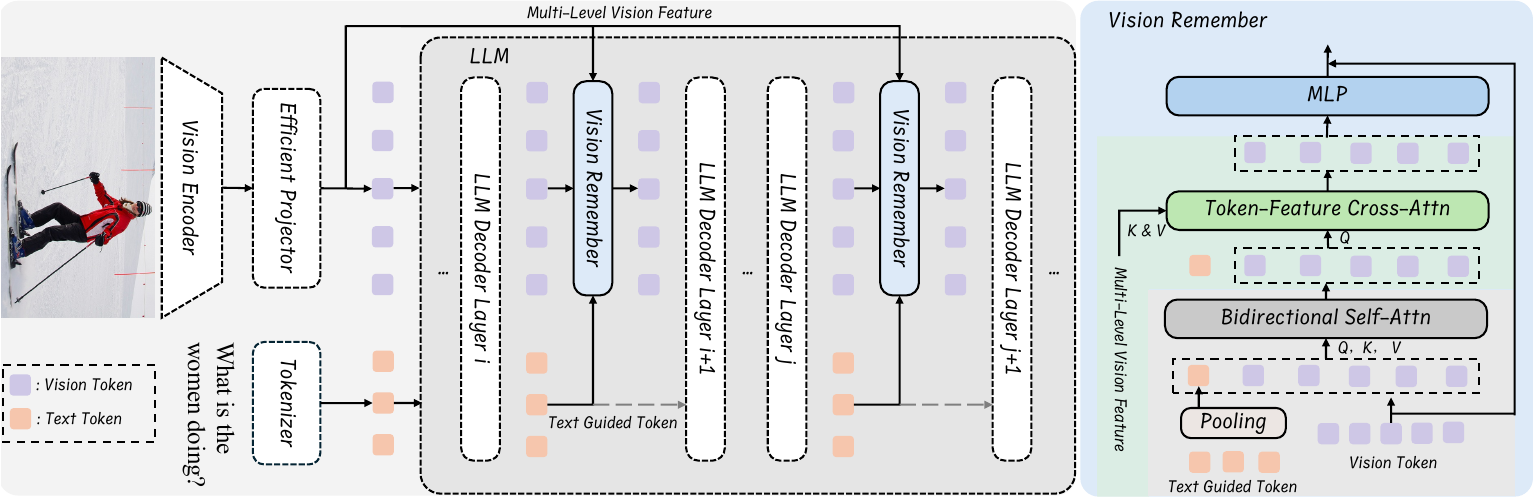}
    \caption{Overview of the proposed \textbf{Vision Remember.} Left part: we insert Vision Remember between the LLM decoder layers to overcome the information bottleneck in token compression and visual cues forgetting in progressive alignment. Right part: Vision Remember consists of two key components: (1) Token-Feature Cross-Attention Layer (shown in the green part) and (2) Token Bidirectional Self-Attention Layer (shown in the gray part).}
    \label{fig:framework}
\end{figure*}

In this section, we first give a brief introduction to the widely used LLaVA series \citep{liu2024llavanext, li2024llavaov, liu2024improved, liu2024visual}, which serves as our baseline. 
We then introduce our proposed Vision Remember, including two key components: Token-Feature Cross-Attention Layer and Token Bidirectional Self-Attention Layer. 
Notably, Vision Remember is not only bound to LLaVA but also can be integrated into other Efficient LVLMs. 
The experiment results on different baselines are shown in Sec.\ref{sec:more_analysis}.

\subsection{Preliminary}
We choose the widely used LLaVA-NeXT \citep{liu2024llavanext} as our baseline, which consists of three components: 1) Vision Encoder, 2) Vision Projector, and 3) Large Language Model.
Vision Encoder, typically a Vision Transformer (ViT) or Convolution Neural Network (CNN) that has been trained on a large amount of data, is primarily used to extract vision features from the input image. 
Then, a 2-layer MLP named Vision Projector is adopted to align the vision features with linguistic space. 
Finally, the text tokens $\textbf{T}_t$ and the vision tokens $\textbf{T}_v$ after alignment are concatenated and fed into an LLM to generate the response $\textbf{R}$ with length $L$ in an auto-regressive manner: 
\begin{equation}
    p(\textbf{R}|\textbf{T}_v,\textbf{T}_t) = \prod_{l=1}^Lp(r_l|\textbf{T}_v,\textbf{T}_t,r_{<l}),
\end{equation}
where $r_l$ indicates the current generated response token and $r_{<l}$ indicates the previous generated response tokens.

\subsection{Token-Feature Cross-Attention Layer}
As mentioned above, we retain the original vision feature and interact with the vision tokens from the LLM decoder layers to compensate for the lost visual information and recover the forgotten visual cues in progressive alignment. The proposed Token-Feature Cross-Attention layer consists of two key designs: (1) multi-level vision feature fusion and (2) local cross attention.

\noindent\textbf{Multi-level Vision Feature.} Many studies have demonstrated that different layers in ViT \citep{dosovitskiy2020vit} exhibit different attention patterns. 
Shallow layers tend to focus on low-level local spatial information, while deeper layers tend to emphasize global semantic features.
Effectively utilizing the multi-level vision features can significantly enhance the LVLM's parsing and understanding capability.
Here, given vision features with spatial shape $W\times H$ from $L$ different layers in encoder, we directly concatenate the vision features from different layers along the feature dimension to form information-rich vision features $F_{v} \in \mathbb{R}^{B\times W\times H\times (L\cdot D_v)}$, where $B$ is batch\_size, $D_v$ is the vision feature dimension.

\begin{figure}[t]
    \centering
    \includegraphics[width=0.75\linewidth]{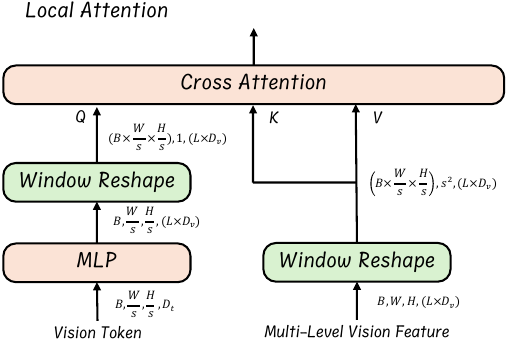}
    \caption{\textbf{Local Cross Attention.} We adopt the local cross attention in Token-Feature Cross-Attention Layer to address the issue of information bottleneck in token compression. A vision token only focuses on a $s\times s$ local region in the multi-level vision feature to improve the computational efficiency and capture the fine-grained spatial information.}
    \label{fig:local}
\end{figure}

\noindent\textbf{Local Cross Attention.} 
To address the issue of information bottleneck in token compression, and better utilize spatial structural information and inductive bias in images, we adopt the local attention mechanism during the token-feature interaction process.
As shown in Fig.\ref{fig:local}, given the vision token $T_{v} \in \mathbb{R}^{B\times \frac{W}{s} \times \frac{H}{s} \times D_t}$ from the LLM decoder layers ($D_t$ is hidden dimension in LLM and $s$ is the downsample ratio of efficient projector), we first expand its dimension by a MLP layer to match the vision features dimension $L\cdot D_v$.
Then following Swin Transformer \citep{liu2021swin}, we divide the vision features $F_{v}$ into $n^2$ local regions with size $s\times s$ in the spatial dimension, where $n=H/s$.
So, we reshape the partitioned vision features $F_v \in \mathbb{R}^{B \times W \times H \times (L\cdot D_v)}$ to $\hat{F_v} \in \mathbb{R}^{  (B\cdot n^2) \times s^2 \times (L\cdot D_v)}$, to serve as key and value.
For vision tokens $T_{v}$, we perform the same window reshape, but the size of the local region is $1\times 1$, serving as the query $\hat{T_{v}}\in \mathbb{R}^{  (B\cdot n^2) \times 1 \times (L\cdot D_v)}$.
In this way, a vision token only performs cross-attention with an $s\times s$ local region, rather than attending globally to all vision features.
Finally, we employ standard cross-attention as follows:
\begin{equation}
    out = Attention(Q=\hat{T_v}, K=\hat{F_v}, V=\hat{F_v}).
\end{equation}

The benefits of using local attention can be summarized as follows: (1) computational efficiency and (2) localized contextual information.
First, local attention reduces the computational complexity compared to traditional global attention mechanisms. By partitioning the vision features into smaller local regions, each vision token only attends to a limited number of vision features, resulting in faster processing and improved efficiency.
Second, local attention allows each vision token to focus on a specific local region of vision features. This attention mechanism helps capture more fine-grained contextual information and spatial relationships within the region, leading to better understanding and representation of the visual content.

\noindent\textbf{Other Attention Mechanisms.} There are three optional interaction mechanisms in Token-Feature Cross-Attention Layer: (1) Local Cross Attention, (2) Deformable Cross Attention \citep{zhu2020deformable, shen2024mome}, and (3) Naive Global Attention \citep{vaswani2017attention}.
For Deformable Cross Attention and Naive Global Attention, we all use vision tokens as query, vision features as key \& value, but the difference is that the former uses deformable attention to deal with multi-level vision features and enhance sparse spatial information.
The comparison between the three interaction mechanisms in Sec.\ref{sec:abalation} shows that Local Cross Attention and Deformable Cross Attention both get positive promotion, and the former gets the best performance.

\subsection{Token Bidirectional Self-Attention Layer}
As mentioned in Sec.\ref{sec:intro}, causal attention mask used in the LLM decoder ensures each token can only attend to preceding tokens in the sequences.
This is naturally suited for language modeling, as textual signals are inherently sequential.
However, visual signals are inherently two-dimensional and encode rich spatial relationships. 
Imposing causal masking during visual token modeling would fundamentally restrict cross-token interactions in visual representations \citep{liu2024vmamba, zhu2024vision, li2024autoregressive}.

In addition to disrupting visual modeling, the causal attention mask also hinders the information flow between vision and language.
In current LVLMs such as LLaVA \cite{zhangllava} and Qwen2.5-VL \cite{bai2025qwen2}, vision tokens are typically prepended to text tokens during sequence concatenation. 
The inherent property of causal attention prevents vision tokens from perceiving subsequent text tokens, effectively causing the model to disregard user prompt inputs when processing vision tokens.
User prompts often contain referential attributes for target objects or foreground elements ($e.g.$, ‘the boy wearing red’ versus ‘the girl wearing blue’). 
Ignoring such input priors prevents the model from distinguishing which vision tokens actually merit attention ($i.e.$, those containing the referenced foreground) during vision token processing.

To address these problems, we introduce Token Bidirectional Self-Attention Layer, which employs self-attention mechanism with full attention.
Given the hidden states in the decoder layer $i$, we first extract text tokens $T_{t} \in \mathbb{R}^{B\times N_{t} \times D_t}$ ($N_{t}$ is the number of text tokens), and then compress them along the sequence dimension with Adaptive Max Pooling to get the text-guided token $\hat{T_{t}} \in \mathbb{R}^{B\times 1 \times D_t}$.
Finally, we concatenate vision tokens $T_{v}$ with the $\hat{T_{t}}$ to enable fully cross-modal interaction through the standard self-attention mechanism.

\subsection{Training.}
Following the common practice, we train the LVLM in multiple phases.

\noindent\textbf{Phase-1: Language-Image Alignment.} 
In this phase, we use the image-caption pairs in the CC-558K dataset \citep{liu2024visual} to train the Vision Projector and Vision Remember, keeping the Vision Encoder and LLM frozen. 
The main purpose of this phase is to align the hidden representation space between the vision and language modalities.

\noindent\textbf{Phase-2: Visual Instruct Tuning.} 
In this phase, we include the LLM in training.
The 779K mixture dataset \citep{liu2024llavanext} is used to enhance the LVLM's ability of visual understanding and instruction following.
To support high-resolution input images, the AnyRes \citep{li2024llavaov} technique is adopted during this phase.

\section{Experiments}
\label{sec:exp}
\subsection{Implementation Details}
We choose LLaVA-NeXT \citep{liu2024llavanext} as baseline, SigLip-Large \citep{zhai2023sigmoid} as Vision Encoder and Qwen2 \citep{qwen2} series as LLM. For fast validation, we all use Qwen2-0.5B if not specified.
The size of each tile of image is resized to $384\times 384$, so the shape of feature map from SigLip-Large-patch16-384 is $24\times 24$, and then 2D Adaptive Average Pooling is employed to compress the spatial resolution to $8\times 8$, resulting in 64 vision tokens per patch ($i.e.$ compression ratio is 1/9).
If not specified, we select layers 7, 15, and 23 from the vision encoder to form multi-level vision features and insert Vision Remember after the first and fourth decoder layers.
We train all models for one epoch, and use the AdamW optimizer with Cosine learning rate schedule.
In phase-1, the learning rate is 1e-3 and the batch size is 256, and in phase-2, the learning rate is 2e-4 and the batch size is 32.
The experiments are conducted on 8 $\times$ Nvidia H20 GPUs.

\subsection{Benchmarks}
We conduct extensive experiments on 11 benchmarks to validate the understanding and parsing capabilities of the proposed method.
The benchmarks can be divided into the following types based on different focus areas:
(1) General Question Answer benchmarks include GQA \citep{gqa}, MME-Perception \citep{mme} and RealWorldQA \citep{rwqa}.
(2) Comprehensive Knowledge Reasoning benchmarks include ScienceQA\_Image \citep{scienceqa}, AI2D \citep{ai2d}, MMMU \citep{mmmu} and MMStar \citep{mmstar}.
(3) OCR\&Chart Parsing benchmarks include ChartQA \citep{chartqa}, DocVQA \citep{docvqa}, TextVQA \citep{textvqa} and OCRBench \citep{ocrbench}.
To compare the accuracy of LVLMs, we take the average score on the whole benchmarks.
Due to the page's limitation, we only report the average scores of these types benchmarks, and the detailed results can be found in Supplementary Material.

\subsection{Main Results}
\noindent\textbf{Accuracy gain with various efficient vision projectors.}
To demonstrate the effectiveness of Vision Remember, we report the performance when combined with various efficient vision projectors \citep{yao2024deco,shen2024mome,chen2024far,chu2023mobilevlm,chu2024mobilevlm}.
Just as Tab.\ref{tab:projector} shows, when different projectors are combined with Vision Remember, the LVLM's ability for visual understanding is all improved.
Specifically, the proposed method can lift the average score of \textit{Adaptive Average Pooling} by +3.0, \textit{PixelShuffle} by +3.2, and \textit{Perceiver Resamplers} by +4.4.
The higher improvements are primarily concentrated on benchmarks including General and OCR\&Chart, which demonstrates that Vision Remember can alleviate the visual information loss and enhance the LVLM’s ability to understand fine-grained visual features and spatial relationships, especially in tasks such as OCR and Chart/Table analysis.
\begin{table}[tp]
\centering
\scriptsize
\setlength{\tabcolsep}{4.5pt}
\renewcommand{\arraystretch}{1.2}
\caption{\textbf{Accuracy gain with various efficient vision projectors.} Performance with proposed \textbf{Vision Remember} (termed as \textit{V.R.}) is marked in \colorbox{blue!7}{blue}. Qwen2-0.5B is used as LLM and the compression ratio is 1/9. Detailed score of each benchmark could be found in \textit{Supplementary Material}. 
$\uparrow$ means that higher is better.
\textit{Our proposed method can improve the LVLM’s ability of visual parsing and understanding when combined with various efficient vision projectors.}}
\label{tab:projector}
\begin{tabular}{c|c|ccc|c}
{{Projectors}} & {V.R.} & {General} & {Knowledge} & {OCR\&Chart} & {\textbf{Average} $\uparrow$} \\ 
\hline
& & 55.2 & 43.4 & 40.4 & \textbf{45.5}\textcolor{gray}{($\Delta0.0$)} \\
\rowcolor{blue!7}
\cellcolor{white}\multirow{-2}{*}{Pooling} & \ding{51} & 56.5 & 43.6 & 47.5 & \textbf{48.5}(\textcolor{red}{+3.0}) \\
\hline
& & 55.4 & 43.5 & 37.4 & \textbf{44.6}\textcolor{gray}{($\Delta0.0$)} \\
\rowcolor{blue!7}
\cellcolor{white}\multirow{-2}{*}{PixelShuffle} & \ding{51} & 55.6 & 43.4 & 46.8 & \textbf{48.0}(\textcolor{red}{+3.4}) \\
\hline
& & 54.1 & 43.7 & 35.4 & \textbf{43.5}\textcolor{gray}{($\Delta0.0$)} \\
\rowcolor{blue!7}
\cellcolor{white}\multirow{-2}{*}{Percevier} & \ding{51} & 55.5 & 43.5 & 47.3 & \textbf{48.1}(\textcolor{red}{+4.6}) \\
\hline
& & 55.7 & 43.1 & 42.2 & \textbf{46.2}\textcolor{gray}{($\Delta0.0$)} \\
\rowcolor{blue!7}
\cellcolor{white}\multirow{-2}{*}{LDPv2} & \ding{51} & 56.1 & 43.6 & 46.9 & \textbf{48.2}(\textcolor{red}{+2.0}) \\
\end{tabular}
\end{table}

\noindent\textbf{Accuracy gain with various compression ratios.}
Tab.\ref{tab:ratio} presents the performance gains with various compression ratios.
We first employ the Adaptive Average Pooling to compress the vision tokens with three ratios: 1/4, 1/9 and 1/16, \textit{i.e} 144, 64 and 36 vision tokens remain in each patch, respectively. 
Then we integrate the proposed Vision Remember and compare the average score on 11 benchmarks.
Specifically, our method achieves performance gains of +2.4, +3.0, and +4.1 at compression ratios of 1/4, 1/9, and 1/16, respectively. 
These results demonstrate that Vision Remember consistently improves performance across varying compression rates, with greater performance gains observed at higher compression ratios.
\begin{table}[tp]
\centering
\scriptsize
\setlength{\tabcolsep}{4.5pt}
\renewcommand{\arraystretch}{1.2}
\caption{\textbf{Accuracy gain with various compression ratios.} Performance with proposed \textbf{Vision Remember} is marked in \colorbox{blue!7}{blue}. \textit{Adaptive Average Pooling} is used in the projector to down-sample the vision tokens and Qwen2-0.5B is used as LLM.
Detailed score of each benchmark could be found in \textit{Supplementary Material}.
$\uparrow$ means that higher is better. 
\textit{The proposed method demonstrates consistent performance improvements across varying compression ratios, with greater performance gains observed at higher compression ratios (\textit{i.e.}, fewer retained tokens).}}
\label{tab:ratio}
\begin{tabular}{c|c|ccc|c}
{{Comp. Ratio}} & {V.R.} & {General} & {Knowledge} & {OCR\&Chart} & {\textbf{Average} $\uparrow$} \\ 
\hline
& & 54.7 & 43.6 & 45.3 & \textbf{47.3}\textcolor{gray}{($\Delta0.0$)} \\
\rowcolor{blue!7}
\cellcolor{white}\multirow{-2}{*}{1/4} & \ding{51} & 56.5 & 44.1 & 50.1 & \textbf{49.7}(\textcolor{red}{+2.4}) \\
\hline
& & 55.2 & 43.4 & 40.4 & \textbf{45.5}\textcolor{gray}{($\Delta0.0$)} \\
\rowcolor{blue!7}
\cellcolor{white}\multirow{-2}{*}{1/9} & \ding{51} & 56.5 & 43.6 & 47.5 & \textbf{48.5}(\textcolor{red}{+3.0}) \\
\hline
& & 53.9 & 43.7 & 35.2 & \textbf{43.3}\textcolor{gray}{($\Delta0.0$)} \\
\rowcolor{blue!7}
\cellcolor{white}\multirow{-2}{*}{1/16} & \ding{51} & 55.9 & 44.1 & 45.2 & \textbf{47.7}(\textcolor{red}{+4.4}) \\
\end{tabular}
\end{table}

\noindent\textbf{Comparison with previous vision feature re-fusion methods.}
Tab.\ref{tab:compare_refusion} presents the accuracy gain comparison with previous vision feature re-fusion methods \cite{meng2024deepstack, cambrian} on the same baseline.
When compared with DeepStack \cite{meng2024deepstack}, we adopt the most widely used implementation in Qwen3-VL, and keep the other setting consistent with ours for fair comparison.
As shown in Tab.\ref{tab:compare_refusion}, the proposed Vision Remember consistently outperforms previous vision feature re-fusion approaches across all evaluated model scales. 
For Qwen2-0.5B as LLM, Vision Remember achieves the highest average score of 48.5, surpassing the baseline by +3.0, SVA Aggregator by +3.4 and DeepStack by +3.9. 
This performance gain is also observed with larger models: for Qwen2-1.5B, Vision Remember improves the average accuracy by +2.9 points over the baseline and outperforms DeepStack by +3.0, and for Qwen2-7B, it achieves an average score of 60.2, which is +2.0 points higher than the baseline and +2.4 than DeepStack and SVA Aggregator.
Notably, Vision Remember demonstrates substantial improvements in the OCR\&Chart category, indicating its superior ability to retain and utilize visual cues and alleviate the visual information loss which is mentioned above. 
\begin{table}[tp]
\centering
\scriptsize
\setlength{\tabcolsep}{4.5pt}
\renewcommand{\arraystretch}{1.2}
\caption{\textbf{Accuracy comparison with previous vision feature re-fusion methods.} Performance with proposed \textbf{Vision Remember} is marked in \colorbox{blue!7}{blue}. \textit{Adaptive Average Pooling} is used in the projector to downsample the vision tokens and the compression ratio is 1/9.
Detailed score of each benchmark could be found in \textit{Supplementary Material}.
\textit{Vision Remember outperforms previous re-fusion methods, indicating its superior ability to retain and utilize visual cues and alleviate the visual information loss.}}
\label{tab:compare_refusion}
\begin{tabular}{c|ccc|c}
{{Methods}} & {General} & {Knowledge} & {OCR\&Chart} & {\textbf{Average} $\uparrow$} \\ 
\hline
\multicolumn{5}{c}{\textit{Qwen2-0.5B as LLM}} \\
\hline
Baseline & 55.2 & 43.4 & 40.4 & \textbf{45.5}\textcolor{gray}{($\Delta0.0$)} \\
\rowcolor{lightgray!20}
SVA Aggregator \cite{cambrian} & 54.4 & 43.6 & 39.4 & \textbf{45.1}(\textcolor{teal}{-0.4}) \\
DeepStack \cite{meng2024deepstack} & 54.0 & 43.5 & 38.7 & \textbf{44.6}(\textcolor{teal}{-0.9}) \\
\rowcolor{blue!7}
\textbf{Vision Remember} & 56.5 & 43.6 & 47.5 & \textbf{48.5}(\textcolor{red}{+3.0}) \\
\hline
\multicolumn{5}{c}{\textit{Qwen2-1.5B as LLM}} \\
\hline
Baseline & 60.2 & 51.2 & 49.8 & \textbf{53.2}\textcolor{gray}{($\Delta0.0$)} \\
\rowcolor{lightgray!20}
SVA Aggregator \cite{cambrian} & 61.0 & 51.0 & 49.8 & \textbf{53.2}(\textcolor{teal}{+0.0}) \\
DeepStack \cite{meng2024deepstack} & 60.6 & 51.3 & 49.3 & \textbf{53.1}(\textcolor{teal}{-0.1}) \\
\rowcolor{blue!7}
\textbf{Vision Remember} & 63.2 & 51.3 & 55.6 &\textbf{56.1}(\textcolor{red}{+2.9}) \\
\hline
\multicolumn{5}{c}{\textit{Qwen2-7B as LLM}} \\
\hline
Baseline & 63.9 & 56.0 & 56.4 & \textbf{58.2}\textcolor{gray}{($\Delta0.0$)} \\
\rowcolor{lightgray!20}
SVA Aggregator \cite{cambrian} & 63.2 & 55.0 & 56.6 & \textbf{57.8}(\textcolor{teal}{-0.4}) \\
DeepStack \cite{meng2024deepstack} & 63.6 & 54.6 & 56.8 & \textbf{57.8}(\textcolor{teal}{-0.4}) \\
\rowcolor{blue!7}
\textbf{Vision Remember} & 66.2 & 56.9 & 58.9 &\textbf{60.2}(\textcolor{red}{+2.0}) \\
\end{tabular}
\end{table}

\noindent\textbf{Comparison with other efficient methods.}
\begin{table}[tp]
\centering
\scriptsize
\renewcommand{\arraystretch}{1.2}
\caption{\textbf{Accuracy comparison with other efficient methods.} We reproduce these methods under the consistent settings (\textit{the same number or comparable total number of vision tokens, the same baseline, and the same training data}). 
The compression ratio is 1/9. 
Performance with proposed \textbf{Vision Remember} is marked in \colorbox{blue!7}{blue}. Performance with  \textit{vision token pruning} is marked in \colorbox{lightgray!20}{gray}. 
Detailed score of each benchmark could be found in \textit{Supplementary Material}. 
\textcolor{teal}{Green} means performance drop compared with our method.
\textit{Our proposed method outperforms previous efficient approaches with a large margin.}
}
\label{tab:com}
\begin{tabular}{c|ccc|c}
{Methods} & {General} & {Knowledge} & {OCR\&Chart} & {\textbf{Average} $\uparrow$} \\  
\hline
\multicolumn{5}{c}{\textit{Qwen2-0.5B as LLM}} \\
\hline
\rowcolor{lightgray!20}
FastV \citep{chen2024image} & 53.4 & 42.6 & 35.2 & \textbf{42.8(\textcolor{teal}{-5.7})}\\
\rowcolor{lightgray!20}
PDrop \citep{xing2024pyramiddrop} & 55.4 & 43.9 & 39.1 & \textbf{45.3(\textcolor{teal}{-3.2})}\\
\rowcolor{lightgray!20}
VisPruner \citep{vispruner} & 55.5 & 43.9 & 42.7 & \textbf{46.6(\textcolor{teal}{-1.9})}\\
DeCo \citep{yao2024deco} & 55.2 & 43.4 & 40.4 & \textbf{45.5(\textcolor{teal}{-3.0})} \\
TokenPacker \citep{li2025tokenpacker} & 55.5 & 43.9 & 40.6 & \textbf{45.8(\textcolor{teal}{-2.7})} \\
\rowcolor{blue!7}
\textbf{Vision Remember} & 56.5 & 43.6 & 47.5 & \textbf{48.5}\textcolor{gray}{($\Delta0.0$)} \\
\hline
\multicolumn{5}{c}{\textit{Qwen2-1.5B as LLM}} \\
\hline
\rowcolor{lightgray!20}
FastV \citep{chen2024image} & 59.0 & 50.7 & 42.3 & \textbf{50.0(\textcolor{teal}{-6.1})} \\
\rowcolor{lightgray!20}
PDrop \citep{xing2024pyramiddrop} & 59.5 & 50.8 & 47.5 & \textbf{52.0(\textcolor{teal}{-4.1})} \\
\rowcolor{lightgray!20}
VisPruner \citep{vispruner} & 60.2 & 51.0 & 52.9 & \textbf{54.1(\textcolor{teal}{-2.0})} \\
DeCo \citep{yao2024deco} & 60.2 & 51.2 & 49.8 & \textbf{53.2(\textcolor{teal}{-2.9})} \\
TokenPacker \citep{li2025tokenpacker} & 60.8 & 50.3 & 48.6 & \textbf{52.5(\textcolor{teal}{-3.6})} \\
\rowcolor{blue!7}
\textbf{Vision Remember} & 63.2 & 51.3 & 55.6 &\textbf{56.1}\textcolor{gray}{($\Delta0.0$)} \\
\hline
\multicolumn{5}{c}{\textit{Qwen2-7B as LLM}} \\
\hline
\rowcolor{lightgray!20}
FastV \citep{chen2024image} & 64.2 & 53.5 & 49.3 & \textbf{54.9(\textcolor{teal}{-5.3})} \\
\rowcolor{lightgray!20}
PDrop \citep{xing2024pyramiddrop} & 64.2 & 54.1 & 53.2 & \textbf{56.5(\textcolor{teal}{-3.7})} \\
\rowcolor{lightgray!20}
VisPruner \citep{vispruner} & 64.6 & 54.5 & 57.4 & \textbf{58.3(\textcolor{teal}{-1.9})} \\
DeCo \citep{yao2024deco} & 63.9 & 56.0 & 56.4 & \textbf{58.2(\textcolor{teal}{-2.0})} \\
TokenPacker \citep{li2025tokenpacker} & 64.0 & 55.8 & 53.8 & \textbf{57.3(\textcolor{teal}{-2.9})} \\
\rowcolor{blue!7}
\textbf{Vision Remember} & 66.2 & 56.9 & 58.9 &\textbf{60.2}\textcolor{gray}{($\Delta0.0$)} \\
\end{tabular}
\end{table}
Tab.\ref{tab:com} presents the performance comparison with other efficient methods, including the pruning-based methods FastV \citep{chen2024image}, PyramidDrop \citep{xing2024pyramiddrop}, VisPruner \citep{vispruner}, and compress-based methods DeCo \citep{yao2024deco}, TokenPacker \citep{li2025tokenpacker}.
For fair comparison, we keep the experiment under consistent settings, including the training data, model size, and compression ratio.
Due to the training-free methods need a pretrained LVLM model, we firstly train a LLaVA-NeXT model without token pruning or compression, and then add the pruning strategies.
Across all tested LLM scales, Vision Remember consistently achieves the highest average accuracy, outperforming competing approaches by a considerable margin in Tab.\ref{tab:com}: 48.5 with Qwen2‑0.5B (+1.9 over VisPruner and +3.3 over TokenPacker), 55.5 with Qwen2‑1.5B (+1.4 over VisPruner and +3.0 over TokenPacker), and 60.2 with Qwen2‑7B (+1.6 over VisPruner and +2.9 over TokenPacker).
The performance boost is most pronounced in the OCR\&Chart category, demonstrating Vision Remember’s strong capability for visual information retention and utilization. 
Notably, VisPruner prunes redundant visual tokens in the vision encoder, while FastV and PyramidDrop perform token pruning within the LLM. 
All these methods rely on attention maps to determine which tokens to retain or drop. 
However, Flash Attention \citep{dao2022flashattention} and Scaled Dot-Product Attention (SDPA)—widely adopted techniques for accelerating attention computation—do not support the output of attention maps by design. 
Consequently, the aforementioned pruning methods cannot be fully integrated with these accelerating techniques at certain layers, leading to significant efficiency bottlenecks. 
We will provide a detailed comparative analysis in Sec.\ref{sec:efficiency}.
Compared with DeCo and TokenPacker, our method not only considers the information bottleneck in token compression, but also recovers the lost visual cues in progressive alignment, thus achieving better performance.

\subsection{Abaltion Study}
\label{sec:abalation}
\begin{table}[t]
    \centering
    \caption{Ablation studies. Qwen2-0.5B is LLM, AnyRes4 is used and the compression ratio is 1/9. Detailed score of each benchmark could be found in \textit{Supplementary Material}. The default setting is marked in \colorbox{blue!7}{blue}.
    }
    \begin{subtable}[t]{0.49\textwidth}
        \centering
        \scriptsize
        \setlength{\tabcolsep}{2.5pt}
        \renewcommand{\arraystretch}{1.2}
        \caption{Ablation study of key components.}
        \label{tab:key}
        \begin{tabular}{c|ccc|c}
        Components & General & Knowledge & OCR\&Chart & \textbf{Average Score} $\uparrow$\\ 
        \hline
        Baseline & 54.6 & 43.7 & 33.4 & \textbf{42.9}\textcolor{gray}{($\Delta0.0$)} \\ \hline
        \rowcolor{lightgray!20}
        + Local Attention & 56.4 & 43.7 & 39.5 &\textbf{45.7}(\textcolor{red}{+2.8}) \\
        + Multi-level Fusion & 55.4 & 44.0 & 42.2 &\textbf{46.3}(\textcolor{red}{+0.6}) \\
        \rowcolor{lightgray!20}
        + Bidirection Interaction & 55.7 & 43.9 & 42.4 &\textbf{46.6}(\textcolor{red}{+0.3}) \\
        \rowcolor{blue!7}
        + Text Token Guidence & 55.9 & 43.9 & 42.5 &\textbf{46.7}(\textcolor{red}{+0.1}) \\
        \end{tabular}
        \vspace{0.5em}
    \end{subtable} 
    \begin{subtable}[t]{0.49\textwidth} 
        \centering
        \scriptsize
        \setlength{\tabcolsep}{2.5pt}
        \renewcommand{\arraystretch}{1.2}
        \caption{Experimental results with various \textbf{interaction methods} in Vision Remember. 
        \textit{Local Attention achieves the best accuracy}.
        }
        \begin{tabular}{c|ccc|c}
            {Interaction Attention} & {General} & {Knowledge} & {OCR\&Chart} & {\textbf{Average Score} $\uparrow$} \\ 
            \hline
            Global Attention & 54.1 & 43.3 & 36.4 & \textbf{43.7} \\
            \rowcolor{lightgray!20}
            Deformable Attention  & 54.8 & 43.0 & 40.0 & \textbf{45.1} \\
            \rowcolor{blue!7}
            Local Attention & 55.9 & 43.9 & 42.5 & \textbf{46.7} \\
        \end{tabular}
        \label{tab:ablabtion_attn}
        \vspace{0.5em}
    \end{subtable}
    \begin{subtable}[t]{0.49\textwidth} 
        \centering
        \scriptsize
        \setlength{\tabcolsep}{2.5pt}
        \renewcommand{\arraystretch}{1.4}
        \caption{Experimental results with \textbf{insertion positions} of Vision Remember.}
        \begin{tabular}{c|ccc|c}
           {Insertion Layers} & {General} & {Knowledge} & {OCR\&Chart} & {\textbf{Average Score} $\uparrow$} \\ 
            \hline
            {1}        & 55.5 & 43.6 & 42.0 & \textbf{46.3} \\
            \rowcolor{blue!7}
            {1,4}      & 55.9 & 43.9 & 42.5 & \textbf{46.7} \\
            {1,4,7}    & 55.9 & 44.0 & 42.5 & \textbf{46.7} \\
        \end{tabular}
        \label{tab:insert}
        \vspace{0.5em}
    \end{subtable}
    \begin{subtable}[t]{0.49\textwidth} 
        \centering
        \scriptsize
        \setlength{\tabcolsep}{3.0pt}
        \renewcommand{\arraystretch}{1.2}
        \caption{Extra \textbf{latency} and \textbf{GPU memory cost} on a NVIDIA A100 GPU.
        \textit{-}.}
        \begin{tabular}{c|ccc|c}
           {Methods} & {TTFT/ms $\downarrow$} & {TPS $\uparrow$} & {GPU Mem./MB $\downarrow$}& {\textbf{Average Score} $\uparrow$} \\ 
            \hline
            Baseline & 57.6 & 45.2 & 3824 & \textbf{42.9} \\
            \rowcolor{blue!7}
            \textbf{+Ours} & 61.2 & 45.2 & 4004 & \textbf{46.7}(\textcolor{red}{+3.8}) \\
        \end{tabular}
        \label{tab:computational}
    \end{subtable}
    \label{tab:ablation}
\end{table}
\textbf{Key Components.}
Tab.\ref{tab:key} presents the results of the ablation study that evaluate the contributions of different key components. 
By incrementally adding Local Attention, Multi-level Fusion, Bidirectional Interaction, and Text guided Token, the results demonstrate performance gains over the baseline. 
The baseline achieves an average score of 42.9, whereas Local Attention boosts the average to 45.7. Introducing Multi-level Fusion further increases the average to 46.3, and integrating Bidirectional Interaction achieves an average score of 46.6. 
Notably, the simultaneous use of all yields the highest average score of 46.7, an improvement of +3.8 over the baseline. 
The results clearly indicate that each module positively contributes to overall performance, and their combined usage provides the most significant enhancement, particularly in complex tasks such as OCR and Chart understanding. 

\noindent\textbf{Interaction Methods in Vision Remember.}
Tab.\ref{tab:ablabtion_attn} investigates the impact of different interaction methods within the Vision Remember.
Due to its focus on all image features without taking into account the visual local context information, Global Attention yields the poorest results (average score 43.7). 
Deformable Attention takes into account local sparse sampling, but learning the offsets can cause the model confusion about reasonable sampling points, leading to suboptimal result (average score 45.1).
Local Attention achieves the best results (average score 46.7).

\noindent\textbf{Insertion Position in LLM.} We also conduct ablation study on the insertion layers of Vision Remember, and the results are reported in the Tab.\ref{tab:insert}.
If we insert vision remember after the first layer, the average score gets +3.4 improvement and yields 46.3
When we insert vision remember after the first and fourth layers, the average score yields 46.7.
Further insertion into subsequent layers leads to the performance saturating without measurable gains in the average metric.
This is because the middle layers are `thinking and reasoning' \citep{wu2024acceleratingmultimodallargelanguage, yu2025multimodalllmssolveimage, basu2024understandinginformationstoragetransfer}, and introducing too many visual features may destroy this pattern.

\noindent\textbf{Extra computational cost of Vision Remember.}
In Tab.\ref{tab:computational}, we report the per-filling latency (\textit{TTFT}), decoding speed (\textit{TPS}) and the GPU memory cost on baseline and Vision Remember. 
After incorporating our method, TTFT increased by 3.6ms and GPU memory increased by 180MB; however, this is virtually imperceptible in real-world use.
In contrast, the average accuracy improved by +3.8, especially in benchmarks such as OCR\&Chart, where the accuracy gain reaches +9.1, which is a significant improvement.

\subsection{More Analysis}
\label{sec:more_analysis}
\begin{table}[htbp]
    \centering
    \caption{Accuracy gain on different baselines. 
    \textit{Vision Remember achieves improvements on two different baselines}.}
    \scriptsize
    \setlength{\tabcolsep}{2.4pt}
    \renewcommand{\arraystretch}{1.2}
    \begin{tabular}{c|c|ccc|c}
        {Baseline} & V.R.& {General} & {Knowledge} & {OCR\&Chart} & {\textbf{Average Score} $\uparrow$} \\ 
        \hline
        & & 58.1 & 51.2 & 59.5 & \textbf{56.1} \\
        \rowcolor{blue!7}
        \cellcolor{white}\multirow{-2}{*}{Qwen2.5-VL \cite{bai2025qwen2}} & \ding{51} & 59.7 & 53.2 & 60.9 & \textbf{57.8}(\textcolor{red}{+1.7}) \\
        \hline
        & & 56.6 & 57.4 & 40.9 & \textbf{50.4} \\
        \rowcolor{blue!7}
        \cellcolor{white}\multirow{-2}{*}{MiniCPM-V \cite{yao2024minicpm}} & \ding{51} & 58.1 & 58.1 & 41.5 & \textbf{51.5}(\textcolor{red}{+1.1}) \\
    \end{tabular}
    \label{tab:baseline}
\end{table}

\noindent\textbf{Accuracy gain on other baselines.}
We also evaluate the proposed method on two different baselines: Qwen2.5-VL-3B \citep{bai2025qwen2} and MiniCPM-V-3B \citep{yao2024minicpm}.
Qwen2.5-VL employs NaViT \citep{dehghani2023patch} as vision encoder and pixelshuffle merger to compress the vision tokens, while MiniCPM-V uses MiniCPM as LLM and Q-Former like perceiver resampler as projector.
Since none of them released their training data, we re-trained the models on the LLaVA-NeXT \citep{liu2024llavanext} training set, and the final results are reported in Tab.\ref{tab:baseline}.
Our proposed method achieves performance improvements across two baselines, proving its effectiveness and robustness.
This experiment also demonstrates that Vision Remember could be considered as a basic component when constructing an Efficient LVLM.

\begin{figure}[t]
    \centering
    \includegraphics[width=1.0\linewidth]{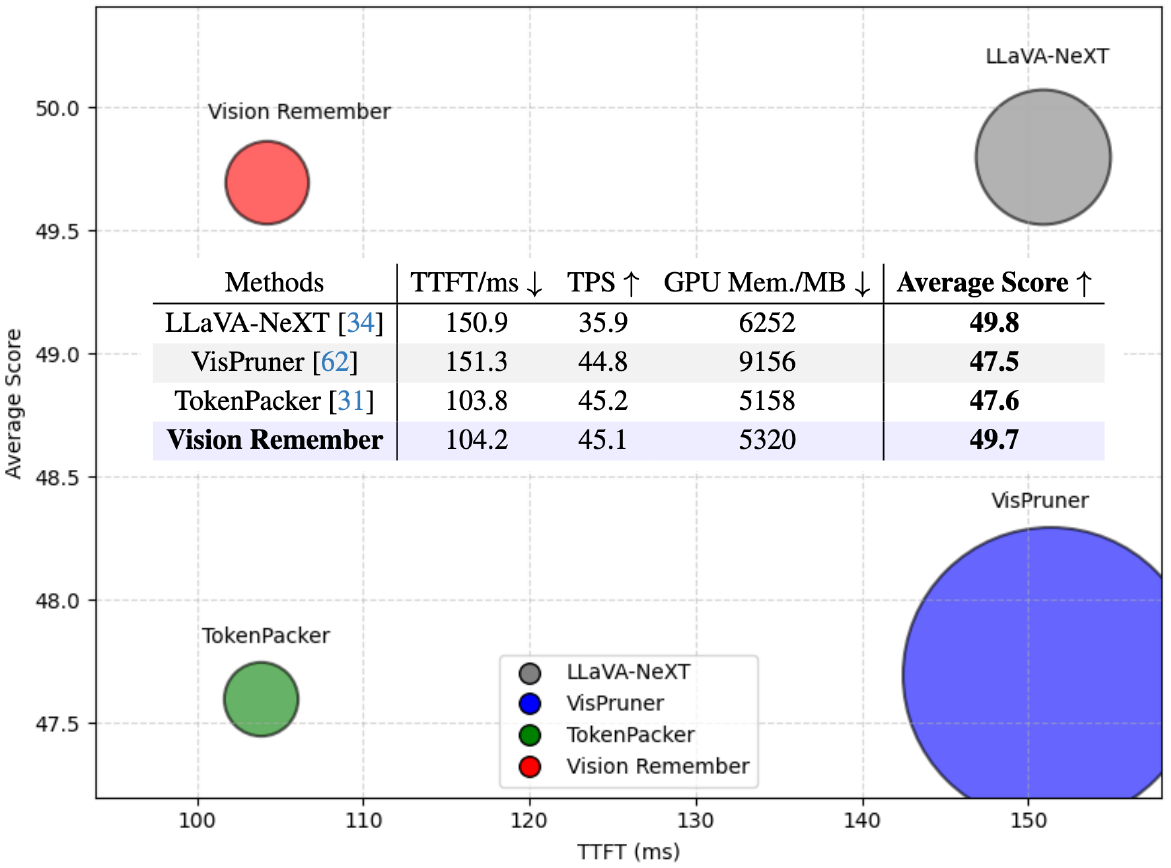}
    \caption{Efficiency comparison on a NVIDIA A100 GPU. The compression ratio is 1/4. The radius of the circles represents the GPU memory used. \textit{Vision Remember achieves the best trade-off on efficiency and accuracy}.}
    \label{fig:efficiency}
\end{figure}
\noindent\textbf{Efficiency analysis.}
\label{sec:efficiency}
Fig.\ref{fig:efficiency} presents the efficiency comparison under compression ratio of 1/4. We chose three metrics, TTFT (Time to First Token), TPS (Tokens per Second) and GPU memory cost, to evaluate the efficiency of the proposed method and others.
TTFT reflects the prefilling stage (limited on computational capacity) latency, and TPS indicates the decoding stage (limited on memory bandwidth) efficiency.
Compared with LLaVA-NeXT \citep{liu2024llavanext}, which does not compress the vision tokens, our method saves 46.7ms (31\%) in the prefilling stage and nearly 1GB GPU memory, and improves TPS to 45.1, while only getting 0.1 (0.2\%) drop on the average score.
Although the Vision Remember module remains inactive during the decoding stage, our method reduces the KV cache length (because of the vision token compression in the prefilling stage) and improves the decoding efficiency.
VisPruner relies on attention maps to select important tokens and cannot be compatible with Flash Attention or SDPA.
Consequently, it cannot accelerate the compute-bound pre-filling phase and gets the highest GPU memory cost.

\section{Conclusion}
\label{sec:conclusion}
In this paper, we investigate the visual information loss in Efficient LVLMs and identify two reasons: Information Bottleneck and Visual Cues Forgetting.
And then we propose Vision Remember to recover the lost visual information with vision feature resampling.
Equipped with Token-Feature Cross-Attention Layer and Token Bidirectional Self-Attention Layer, the proposed method captures more fine-grained contextual information and spatial relationships, enhancing the capability of visual parsing and understanding.
Comprehensive experiments validate the effectiveness of the proposed method when combined with various efficient vision projectors and LVLMs.
We hope our work can promote community interest in Efficient LVLMs.



\end{document}